\documentclass[aps,prx,showkeys,reprint]{revtex4-2}

\usepackage[utf8]{inputenc}
\usepackage[T1]{fontenc}
\usepackage{amsmath}
\usepackage{amssymb}
\usepackage{amsfonts}
\usepackage{upgreek}
\usepackage{mathtools}
\usepackage{float}
\usepackage{graphicx}
\usepackage{tabularx}
\usepackage[labelfont=bf]{caption}

\usepackage[hidelinks]{hyperref}

\begin{document}

\title{CoRe Optimizer: An All-in-One Solution for Machine Learning}

\author{Marco Eckhoff}
\email{eckhoffm@ethz.ch}
\author{Markus Reiher}
\email{mreiher@ethz.ch}
\affiliation{ETH Zurich, Department of Chemistry and Applied Biosciences, Vladimir-Prelog-Weg 2, 8093 Zurich, Switzerland.}

\date{January 17, 2024}

\begin{abstract}

The optimization algorithm and its hyperparameters can significantly affect the training speed and resulting model accuracy in machine learning applications. The wish list for an ideal optimizer includes fast and smooth convergence to low error, low computational demand, and general applicability. Our recently introduced continual resilient (CoRe) optimizer has shown superior performance compared to other state-of-the-art first-order gradient-based optimizers for training lifelong machine learning potentials. In this work we provide an extensive performance comparison of the CoRe optimizer and nine other optimization algorithms including the Adam optimizer and resilient backpropagation (RPROP) for diverse machine learning tasks. We analyze the influence of different hyperparameters and provide generally applicable values. The CoRe optimizer yields best or competitive performance in every investigated application, while only one hyperparameter needs to be changed depending on mini-batch or batch learning.

\end{abstract}

\keywords{Continual Resilient (CoRe) Optimizer, Adam, RPROP, AdaMax, RMSprop, AdaGrad, AdaDelta, NAG, Momentum, SGD}

\maketitle

\section{Introduction}

Machine learning (ML) is a part of the general field of artificial intelligence. ML is employed in a wide range of applications such as computer vision, natural language processing, and speech recognition \cite{Bishop2006, Russell2021}. It involves statistical models whose performance on tasks can be improved by learning from sample data or past experience. ML models include very many parameters, the so-called weights. In the learning process, these weights are optimized according to a performance measure. To evaluate this measure, training data or experience are required. In supervised learning, the model is trained on labeled data to obtain a function that maps the data to its label as in classification and regression tasks. By contrast, in unsupervised learning unlabeled data is trained for categorization. In addition, in reinforcement learning the model is trained through trial and error aiming to maximize its reward. Hence, ML models predict tasks only based on a learned pattern of the data and do not require explicit program instructions for predictions.

The performance measure can be a loss function (also called cost function) that needs to be minimized \cite{Goodfellow2016}. This loss function is usually a sum over contributions from the training data points. Instead of calculating it simultaneously for the full training data set (deterministic or batch learning), a \text{(semi-)randomly} chosen subset of the training data is often employed (stochastic or mini-batch learning). This approach can accelerate the convergence with respect to the total computation time because the loss-function accuracy increase is sub-linear for larger batch sizes. To update the weights of the ML model, first-order gradient-based iterative optimization schemes are dominating the field, since the memory demand and computation time per step of second-order optimizers is often too high. In general, the optimization aims at a loss function's local minimum as a function of the model's weights because it is sufficient for most ML applications to find weight values with low loss rather than the global minimum.

The optimization algorithm can crucially determine the training speed and final performance of ML models \cite{Sun2019}. Therefore, the development of advanced optimizers is an active field of research which can significantly impact the accuracy of predictions in all ML applications. The simplest form of stochastic first-order minimization for high-dimensional parameter spaces is stochastic gradient decent (SGD) \cite{Robbins1951}. In SGD, the negative gradient of the loss function with respect to each weight is multiplied by a constant learning rate and the product is subtracted from the respective weight in each update. The loss function gradient is adapted in stochastic gradient decent with momentum (Momentum) \cite{Polyak1964} and Nesterov accelerated gradient (NAG) \cite{Nesterov1983, Sutskever2013}. These methods aim to improve convergence by a momentum in the weight updates, as the gradients are based on stochastic estimates. In a different fashion, adaptive gradient (AdaGrad) \cite{Duchi2011}, adaptive delta (AdaDelta) \cite{Zeiler2012}, and root mean square propagation (RMSprop) \cite{Hinton2012} apply the ordinary loss function gradient combined with a weight-specific, adapted learning rate. Adaptive moment estimation (Adam) \cite{Kingma2015}, adaptive moment estimation with infinity norm (AdaMax) \cite{Kingma2015}, and our recently developed continual resilient (CoRe) optimizer \cite{Eckhoff2023} combine momentum with individually adapted learning rates. In resilient backpropagation (RPROP) \cite{Riedmiller1993, Riedmiller1994} only the sign of the loss function gradient is employed with individually adapted learning rates.

Apart from these optimizers, which are applied in this work, many more optimizers have been developed for ML applications in recent years. For example, the modification of the first moment estimation of Adam yields Nesterov-accelerated adaptive moment estimation (NAdam) \cite{Dozat2016} and the modification of the second moment AMSGrad \cite{Reddi2018}. Nesterov momentum is also employed in adaptive Nesterov momentum (Adan) \cite{Xie2023}. Moreover, AdaFactor \cite{Shazeer2018}, AdaBound \cite{Luo2019}, AdaBelief \cite{Zhuang2020}, AdamW \cite{Loshchilov2019}, PAdam \cite{Chen2020}, RAdam \cite{Liu2020}, AdamP \cite{Heo2021}, Lamb \cite{You2020}, Gravity \cite{Bahrami2021}, and Lion \cite{Chen2023} are further examples of the large zoo of optimizers. They often represent incremental improvements of parent algorithms. We note that these optimizers can be used in applications beyond ML as well. Furthermore, second-order optimizers have been proposed such as adaptive estimates of the Hessian \text{(AdaHessian)} \cite{Yao2021} and second-order clipped stochastic optimization (Sophia) \cite{Liu2023}. To acquire an overview of the performance differences among these optimizers, extensive benchmarks are required \cite{Schneider2019, Choi2020, Schmidt2021}. Statistical averaging and uncertainty quantification are indispensable in these benchmarks for validation.

To ease the burden on an ML practitioner in the optimizer choice, an optimizer is desired which performs well on diverse ML tasks. Moreover, a generally applicable set of optimizer hyperparameters is required which works out-of-the-box avoiding time consuming hyperparameter tuning. At most, a single intuitive hyperparameter may require to be adapted coarsely, while its value needs to be easy to estimate. Furthermore, the ideal optimizer features fast and smooth convergence to high accuracy with low computational burden. 

The Adam optimizer is not an obviously superior, but viable choice for many ML tasks \cite{Schmidt2021}. Therefore, Adam became the most frequently applied optimizer with adaptive learning rates. Since our CoRe optimizer has outperformed Adam on the task of training a lifelong machine learning potential (lMLP) \cite{Eckhoff2023}, it is obvious to assess its performance on diverse ML tasks and compare the outcome with that of various aforementioned optimizers. Such a broad performance evaluation further allows us to obtain generally valid hyperparameters for the CoRe optimizer to obtain an all-in-one solution.

As a benchmark, we examine a set of fast running ML tasks provided in PyTorch \cite{Paszke2019}. The benchmark set spans the range from small mini-batch learning to full batch learning as well as reinforcement learning. Moreover, it includes different tasks, models, and data sets to enable a broad comparison of different optimizers. First, for the MNIST handwritten digits \cite{Deng2012} and Fashion-MNIST \cite{Xiao2017} data sets we run mini-batch learning to do variational auto-encoding (AED and ADF) \cite{Kingma2014} and image classification (ICD and ICF). The latter is done by convolutional neural networks \cite{Fukushima1980} with rectified linear units (ReLU) \cite{Agarap2019}, dropout \cite{Srivastava2014}, max pooling \cite{Scherer2010}, and softmax. Second, for the cart-pole problem \cite{Barto1983} we perform naive reinforcement learning (NR) with a feed-forward linear neural network \cite{Rosenblatt1958}, dropout, ReLU, and softmax and reinforcement learning by an actor-critic algorithm (RA) \cite{Konda1999}. Third, for the BSD300 data set \cite{Martin2001} we carry out single image super-resolution (SR) with upscale factor four by sub-pixel convolutional neural networks \cite{Shi2016} employing relatively large mini-batches. Fourth, we run batch learning of the Cora data set \cite{Sen2008} for semi-supervised classification (SS) with graph convolutional networks \cite{Kipf2017} and dropout as well as of a sine wave for time sequence prediction (TS) with a long short-term memory (LSTM) cell \cite{Hochreiter1997}.

In addition, we evaluate the optimizers in training of a machine learning potential \cite{Behler2016, Bartok2017, Deringer2019, Noe2020, Westermayr2021, Kaeser2023}, i.e., a regression task. A machine learning potential is a representation of the potential energy surface of a chemical system. It can be employed in atomistic simulations to calculate chemical properties and reactivity. One method example among many others is a high-dimensional neural network potential \cite{Behler2007, Behler2017} which takes as input the chemical element types and atomic coordinates and in required cases atomic charges and spins \cite{Behler2021, Eckhoff2020b, Eckhoff2021a} to calculate the energy and atomic forces of systems ranging from organic molecules over liquids to inorganic materials including multi-component systems such as interfaces \cite{Eckhoff2019, Eckhoff2020a, Eckhoff2021b, Eckhoff2023}. In this work, we repeat the stationary learning of an lMLP based on an ensemble of ten high-dimensional neural network potentials, which employ element-embracing atom-centered symmetry functions as descriptors \cite{Eckhoff2023}. The lMLP is trained on 8600 S$_\text{N}$2 reaction systems with lifelong adaptive data selection.

This work is organized as follows: In Section \ref{sec:Methods}, we summarize the applied optimization algorithms, and in Section \ref{sec:Computational_Details}, we compile the computational details. In Section \ref{sec:Results_and_Discussion}, we analyze the resulting training speed and final accuracy for the PyTorch ML task examples and lMLPs. This work ends with a conclusion in Section \ref{sec:Conclusion}.

\section{Methods}\label{sec:Methods}

\subsection{Continual Resilient (CoRe) Optimizer}

The CoRe optimizer \cite{Eckhoff2023} is a first-order gradient-based optimizer for stochastic and deterministic iterative optimizations. It adapts the learning rates individually for each weight $w_\xi$ depending on the optimization progress. These learning rate adjustments are inspired by the Adam optimizer \cite{Kingma2015}, RPROP \cite{Riedmiller1993, Riedmiller1994}, and the synaptic intelligence method \cite{Zenke2017}.

Exponential moving averages of the loss function gradient and its square,
\begin{align}
g_\xi^\tau&=\beta_1^\tau\cdot g_\xi^{\tau-1}+\left(1-\beta_1^\tau\right)\frac{\partial L^t}{\partial w_\xi^{t-1}}\ ,\label{eq:g}\\
h_\xi^\tau&=\beta_2\cdot h_\xi^{\tau-1}+\left(1-\beta_2\right)\left(\frac{\partial L^t}{\partial w_\xi^{t-1}}\right)^2\ ,\label{eq:h}
\end{align}
with decay rates $\beta_1^\tau,\beta_2\in[0,1)$, are employed in minimization in analogy to the Adam optimizer. For maximization, the sign of the loss function gradient in Equation (\ref{eq:g}) has to be inverted. In the CoRe optimizer, $\beta_1$ is a function of the individual weight update counter $\tau$,
\begin{align}
\beta_1^\tau=\beta_1^\mathrm{b}+\left(\beta_1^\mathrm{a}-\beta_1^\mathrm{b}\right)\exp\left[-\left(\frac{\tau-1}{\beta_1^\mathrm{c}}\right)^2\right]\ ,
\end{align}
whereby $\tau$ can vary from the counter of gradient calculations $t$ if some optimization steps do not update every weight. The initial decay $\beta_1^\mathrm{a}\in[0,1)$ is converted by a Gaussian with width $\beta_1^\mathrm{c}>0$ to the final decay $\beta_1^\mathrm{b}\in[0,1)$. The smaller $\beta_1^\tau$, the higher is the dependence on the current gradient, while a larger $\beta_1^\tau$ leads to a slower decay of previous gradient contributions.

The Adam-like adaption of the weight-specific learning rates,
\begin{align}
u_\xi^\tau=\frac{g_\xi^\tau}{1-\left(\beta_1^\tau\right)^\tau}\left\{\left[\frac{h_\xi^\tau}{1-\left(\beta_2\right)^\tau}\right]^{\tfrac{1}{2}}+\epsilon\right\}^{-1}\ ,\label{eq:u}
\end{align}
employs the quotient of the moving averages $g_\xi^\tau$ and $(h_\xi^\tau)^{\tfrac{1}{2}}$, which are corrected with respect to their initialization bias toward zero ($g_\xi^0,h_\xi^0=0$). For numerical stability $\epsilon\gtrapprox0$ is added in the denominator. This quotient is invariant to gradient rescaling and introduces a form of step size annealing. Therefore, $u_\xi^\tau$ changes from $\pm1$ in the first optimization step $\tau=1$ toward zero in well-behaving optimizations. 

The plasticity factor,
\begin{align}
P_\xi^\tau=\begin{cases}0&\hspace{-0.1cm}\begin{array}{l}\mathrm{for}\ \tau>t_\mathrm{hist}\\
\land\ S_\xi^{\tau-1}\ \mathrm{top}\text{-}n_{\mathrm{frozen},\chi}\ \mathrm{in}\ \mathbf{S}_\chi^{\tau-1}\end{array}\\
1&\hspace{-0.1cm}\begin{array}{l}\mathrm{otherwise}\end{array}\end{cases}\ ,\label{eq:P}
\end{align}
aims to improve the stability-plasticity balance by regularization in the weight updates. Therefore, weight groups $\chi$ are specified---for example, a layer in a neural network---and the weight-specific importance scores $\mathbf{S}_\chi^{\tau-1}$ (see Equation (\ref{eq:S}) below) are compared within these groups. When $\tau>t_\mathrm{hist}>0$, $P_\xi^\tau$ can freeze the weights with the $n_{\mathrm{frozen},\chi}\geq0$ highest importance scores in their group in update $\tau$ to mitigate forgetting of previous knowledge.

The RPROP-like learning rate adaption,
\begin{align}
s_\xi^\tau=\begin{cases}\mathrm{min}\left(\eta_+\cdot s_\xi^{\tau-1},s_\mathrm{max}\right)&\mathrm{for}\ g_\xi^{\tau-1}\cdot g_\xi^\tau\cdot P_\xi^\tau>0\\
\mathrm{max}\left(\eta_-\cdot s_\xi^{\tau-1},s_\mathrm{min}\right)&\mathrm{for}\ g_\xi^{\tau-1}\cdot g_\xi^\tau\cdot P_\xi^\tau<0\\s_\xi^{\tau-1}&\mathrm{for}\ g_\xi^{\tau-1}\cdot g_\xi^\tau\cdot P_\xi^\tau=0\end{cases}\ ,\label{eq:s}
\end{align}
depends only on the sign of the gradient moving average $g_\xi^\tau$ and not on its magnitude leading to a robust optimization. Sign inversions from $g_\xi^{\tau-1}$ to $g_\xi^\tau$ often signalize a jump over a minimum in the previous update. Hence, the step size $s_\xi^{\tau-1}$ is reduced by the decrease factor $\eta_-\in(0,1]$ in this case, while it is enlarged by the increase factor $\eta_+\geq1$ for constant signs to speed up convergence. The updated step size $s_\xi^\tau$ is bounded by the minimal and maximal step sizes $s_\mathrm{min},s_\mathrm{max}>0$. For $g_\xi^{\tau-1}\cdot g_\xi^\tau\cdot P_\xi^\tau=0$, the step size update is omitted. The initial step size $s_\xi^0=s_\xi^1$ is a hyperparameter of the optimization.

The weight decay,
\begin{align}
w_\xi^t=\left(1-d_\chi\cdot\left|u_\xi^\tau\right|\cdot P_\xi^\tau\cdot s_\xi^\tau\right)w_\xi^{t-1}-u_\xi^\tau\cdot P_\xi^\tau\cdot s_\xi^\tau\ ,
\end{align}
with group-specific hyperparameter $d_\chi\in[0,(s_\mathrm{max})^{-1})$, targets to reduce the overfitting risk by prevention of strong weight in- or decreases. It is proportional to the product of $d_\chi$ and the absolute weight update $|u_\xi^\tau|\cdot P_\xi^\tau\cdot s_\xi^\tau$, i.e., the more stable the weight value the less it is affected by the weight decay. Subsequently, the signed weight update $u_\xi^\tau\cdot P_\xi^\tau\cdot s_\xi^\tau$ is subtracted to obtain the updated weight $w_\xi^t$. The weight values are therefore bound between $-(d_\chi)^{-1}$ and $(d_\chi)^{-1}$ in well-behaving optimizations, i.e., $u_\xi^\tau\leq\pm1$.

The importance score value,
\begin{align}
S_\xi^\tau=\begin{cases}\hspace{-0.1cm}\begin{array}{l}S_\xi^{\tau-1}+\left(t_\mathrm{hist}\right)^{-1}g_\xi^\tau\cdot u_\xi^\tau\cdot P_\xi^\tau\cdot s_\xi^\tau\end{array}&\hspace{-0.1cm}\mathrm{for}\ \tau\leq t_\mathrm{hist}\vspace{0.075cm}\\
\hspace{-0.1cm}\begin{array}{l}\left[1-\left(t_\mathrm{hist}\right)^{-1}\right]S_\xi^{\tau-1}\\
+\left(t_\mathrm{hist}\right)^{-1}g_\xi^\tau\cdot u_\xi^\tau\cdot P_\xi^\tau\cdot s_\xi^\tau\end{array}&\hspace{-0.1cm}\mathrm{otherwise}\end{cases}\ ,\label{eq:S}
\end{align}
ranks the weight importance by taking into account weight-specific contributions to previously estimated loss function decreases. This ansatz is inspired by the synaptic intelligence method. The importance scores enable to identify the most important weights in previous updates, which can be frozen by the plasticity factors (Equation (\ref{eq:P})) in following updates to improve the stability-plasticity balance. The product of gradient moving average and signed weight update is employed to estimate the loss function decrease. Since the weight update sign is not inverted, the higher positive the importance score, the larger is the loss function decrease. Starting with $S_\xi^0=0$, the mean of $g_\xi^\tau\cdot u_\xi^\tau\cdot P_\xi^\tau\cdot s_\xi^\tau$ over $\tau\leq t_\mathrm{hist}$ is calculated. For $\tau>t_\mathrm{hist}$, the importance score is determined as exponential moving average with decay $1-(t_\mathrm{hist})^{-1}$.

We note that the relative large number of hyperparameters in the CoRe optimizer is, on the one hand, an advantage to obtain good results even in very difficult or edge cases. On the other hand, the hyperparameter tuning is more complicated. However, a set of generally applicable values, which are provided in this work, can overcome this drawback.

\subsection{SGD}

SGD \cite{Robbins1951} subtracts the product of a constant learning rate $\gamma$ and the loss function gradient from the weights $w_\xi^{t-1}$ in the weight updates,
\begin{align}
w_\xi^t=w_\xi^{t-1}-\gamma G_\xi^\tau\ ,\label{eq:w}
\end{align}
with
\begin{align}
G_\xi^\tau=\frac{\partial L^t}{\partial w_\xi^{t-1}}\ .
\end{align}

\subsection{Momentum}

An additional momentum (Momentum) \cite{Polyak1964} can be introduced in SGD by replacing $G_\xi^\tau$ in Equation (\ref{eq:w}) by
\begin{align}
m_\xi^\tau=\mu\cdot m_\xi^{\tau-1}+G_\xi^\tau\ ,
\end{align}
with the momentum factor $\mu$ and $m_\xi^1=G_\xi^1$ \cite{Paszke2019}.

\subsection{NAG}

NAG \cite{Nesterov1983, Sutskever2013} is SGD with Nesterov momentum, i.e, $G_\xi^\tau$ in Equation (\ref{eq:w}) is substituted by
\begin{align}
n_\xi^\tau=\mu\cdot m_\xi^{\tau}+G_\xi^\tau\ .
\end{align}

\subsection{Adam}

The algorithm of the Adam optimizer \cite{Kingma2015} is given by Equations (\ref{eq:g}) (with constant $\beta_1$), (\ref{eq:h}), (\ref{eq:u}), and (\ref{eq:w}), whereby $G_\xi^\tau$ in Equation (\ref{eq:w}) is replaced by $u_\xi^\tau$. In comparison to the CoRe optimizer, Adam misses the $\tau$ dependence of the decay rate $\beta_1$, the plasticity factors $P_\xi^\tau$, the RPROP-like learning rate adaption $s_\xi^\tau$, and the weight decay. The latter can be introduced in Adam as well as in many other optimizers also by adding $d\cdot w_\xi^{t-1}$ to the loss function gradient as second operation of an optimization iteration after the possible sign inversion for maximization. A further alternative is to subtract instead $d\cdot\gamma\cdot w_\xi^{t-1}$ from $w_\xi^{t-1}$ as in AdamW \cite{Loshchilov2019}.

\subsection{AdaMax}

The difference of the AdaMax optimizer \cite{Kingma2015} compared to Adam is that the term in curly brackets in Equation (\ref{eq:u}) is replaced by the infinity norm,
\begin{align}
k_\xi^\tau=\mathrm{max}\left(\beta_2\cdot k_\xi^{\tau-1},\left|G_\xi^\tau+\epsilon\right|\right)\ ,
\end{align}
with $k_\xi^0=0$.

\subsection{RMSprop}

In RMSprop \cite{Hinton2012} the loss function gradient is divided by the moving average of its magnitude,
\begin{align}
w_\xi^t=w_\xi^{t-1}-\gamma G_\xi^\tau \left[\left(h_\xi^\tau\right)^{\tfrac{1}{2}}+\epsilon\right]^{-1}\ .\label{eq:w_RMSprop}
\end{align}
Hence, the difference to the Adam optimizer is that the loss function gradient $G_\xi^\tau$ is applied instead of the gradient moving average $g_\xi^\tau$ and the initialization bias correction is omitted.

\subsection{AdaGrad}

The AdaGrad optimizer \cite{Duchi2011} differs from RMSprop by the replacement of $h_\xi^\tau$ in Equation (\ref{eq:w_RMSprop}) by
\begin{align}
b_\xi^\tau=b_\xi^{\tau-1}+\left(G_\xi^\tau\right)^2\ ,
\end{align}
with $b_\xi^0=0$.

\subsection{AdaDelta}

The adaptive learning rate in the AdaDelta optimizer \cite{Zeiler2012} is established by
\begin{align}
w_\xi^t=w_\xi^{t-1}-\gamma G_\xi^\tau\left(\frac{l_\xi^{\tau-1}+\epsilon}{h_\xi^\tau+\epsilon}\right)^{\tfrac{1}{2}}\ ,
\end{align}
with
\begin{align}
l_\xi^\tau=\beta_2\cdot l_\xi^{\tau-1}+\left(1-\beta_2\right)\frac{l_\xi^{\tau-1}+\epsilon}{h_\xi^\tau+\epsilon}\left(G_\xi^\tau\right)^2
\end{align}
and $l_\xi^0=0$. Hence, in comparison to the RMSprop algorithm the factor $\left(l_\xi^{\tau-1}+\epsilon\right)^{\tfrac{1}{2}}$ is applied additionally in the weight update and the order of adding $\epsilon$ to $h_\xi^\tau$ and taking the square root is inverted.

\subsection{RPROP}

RPROP \cite{Riedmiller1993, Riedmiller1994} is based on Equation (\ref{eq:s}), whereby $G_\xi^{\tau-1}\cdot G_\xi^\tau$ is employed instead of $g_\xi^{\tau-1}\cdot g_\xi^\tau\cdot P_\xi^\tau$. In addition, a backtracking weight step is applied by setting $G_\xi^\tau=0$ when $G_\xi^{\tau-1}\cdot G_\xi^\tau<0$. The weight update is given by
\begin{align}
w_\xi^t=w_\xi^{t-1}-s_\xi^\tau\cdot\mathrm{sgn}\left(G_\xi^\tau\right)\ .
\end{align}

\section{Computational Details}\label{sec:Computational_Details}

The PyTorch ML task examples \cite{PyTorchExamples2023} were solely modified to embed them in the extensive benchmark without touching the ML models and trainings. The only exception was the removal of the learning rate scheduler in ICD and ICF to assess exclusively the performance of the optimizer. The tasks performed originally only on the MNIST data set (AED and ICD) were also carried out for the Fashion-MNIST data set (AEF and ICF). The batch sizes of the ML tasks AED, AEF, ICD, and ICF were 64 of in total 60000 training data points to obtain test cases for small mini-batch learning (64 was the default value in the ICD PyTorch ML task example). The batch size of SR was 10 of 200 training data points to get an example of a batch size which is a rather large fraction of the total number of data points ($5\%$). The employed scripts with all details on the models, trainings, and error definitions are available on Zenodo \cite{Eckhoff2023b} alongside the compiled raw results as well as plot and analysis scripts. Moreover, this repository as well as the Zenodo repository \cite{Eckhoff2024} contain the CoRe optimizer software, which is compatible to use with \text{PyTorch}. In addition, the lMLP software \cite{Eckhoff2023a} was extended to integrate all optimizers and is also available in the Zenodo repository \cite{Eckhoff2023b} alongside lMLP results as well as model and training details. The latter were taken over from Reference [\citenum{Eckhoff2023}]. The lMLP training employed lifelong adaptive data selection and a fit fraction per epoch of $10\%$ of all 7740 training structures.

Each ML task was performed for each optimizer setting with 20 different sets of random numbers. For reinforcement learning (NR and RA) even 100 different sets of random numbers were employed as the fluctuations in the respective results were the largest. These sets were the same for each optimizer and they ensured differently initialized weights (and different selection of training and test data). The mean test set error $E_i^\mathrm{test}$ and its standard deviation $\Delta E_i^\mathrm{test}$ of ML task $i$ were calculated for each set as a function of the training epoch $n_\mathrm{epoch}$ to evaluate convergence. To determine the final accuracy, for the minimal test set error in each of the 20 trainings the mean $E_i^\mathrm{test,min}$ and standard deviation $\Delta E_i^\mathrm{test,min}$ were calculated, i.e., early stopping was applied. For reinforcement learning (NR and RA) the mean number of training episodes until a reward of 475 \cite{Brockman2016} was taken to quantify $E_i^\mathrm{test}$. The maximum number of training episodes was 2500, which was also used as error of unsuccessful trainings. For AEF 7 of 20 Momentum$^*$ trainings, 8 of 20 NAG trainings, and 12 of 20 NAG$^*$ trainings failed even for the best learning rate value. These trainings were penalized with a constant error of 1000. For lMLPs the total test loss according to Equation (10) in Reference [\citenum{Eckhoff2023}] determined the training epoch with minimal error. In this way, the mean squared error of the energies was weighted with a factor $q^2=10.9^2$ in the loss function, while that of the atomic force components was not scaled. We evaluated the mean error based on the errors of all 20 lMLPs in each of the 20 training epochs where an individual lMLP showed minimal error, i.e., 400 error values were included. In this way, the error was still calculated from advanced training states, while it was also sensitive to the smoothness of the training processes as early stopping is difficult to apply in practise in lifelong machine learning.

To compare the final accuracy among different optimizers $k$ for ML task $i$, the inverse of the minimum test set error $E_{i,k}^\mathrm{test,min}$ relative to the result of best performing optimizer in ML task $i$ was calculated,
\begin{align}
A_i(k)=\dfrac{\min_k\left(E_{i,k}^\mathrm{test,min}\right)}{E_{i,k}^\mathrm{test,min}}\ .\label{eq:A_i}
\end{align}
The uncertainty of the accuracy score was calculated from an error propagation based on the test set error's standard deviation $\Delta E_{i,k}^\mathrm{test}$,
\begin{align}
\Delta A_i(k)=\dfrac{\min_k\left(E_{i,k}^\mathrm{test,min}\right)}{\left(E_{i,k}^\mathrm{test,min}\right)^2}\Delta E_{i,k}^\mathrm{test,min}\ .\label{eq:Delta_A_i}
\end{align}
For comparison of different optimizers $k$ with regard to the overall accuracy, the arithmetic mean $\overline{A}(k)$ over all $N_\mathrm{tasks}$ ML task accuracy scores $A_i(k)$ was calculated,
\begin{align}
\overline{A}(k)=\dfrac{1}{N_\mathrm{tasks}}\sum_i^{N_\mathrm{tasks}}A_i(k)\ .\label{eq:A}
\end{align}
Its uncertainty was determined by propagating the errors of the independent variables $A_i(k)$,
\begin{align}
\Delta\overline{A}(k)=\dfrac{1}{N_\mathrm{tasks}}\left\{\sum_i^{N_\mathrm{tasks}}\left[\Delta A_i(k)\right]^2\right\}^{\tfrac{1}{2}}\ .\label{eq:Delta_A}
\end{align}

The PyTorch version 2.0.0 \cite{Paszke2019} and its default settings were applied for the optimizers AdaDelta, \text{AdaGrad}, Adam, AdaMax, Momentum, NAG, RMSprop, RPROP, and SGD (see Tables S1 and S3 in the Supporting Information for all hyperparameter values). The momentum factor in Momentum and NAG was $\mu=0.9$. In addition, scans of the performance determining hyperparameters $\beta_1$, $\beta_2$, $\mu$, $\eta_-$, and $\eta_+$ were carried out for the PyTorch ML task examples in order to find their optimal values for this set of ML tasks. If the default values turned out to be the best ones, the second best choice was applied. The optimizers employing these modified hyperparameters (see Tables S1 and S3 in the Supporting Information) are marked with an asterisk ($^*$). Weight decay was by default only applied in the CoRe optimizer. The learning rates $s_\xi^0$ of RPROP, RPROP$^*$, and the CoRe optimizer were set to $10^{-3}$. For the learning rate $\gamma$ of the other optimizers and the maximal step size $s_\mathrm{max}$ of RPROP, RPROP$^*$, and the CoRe optimizer, the values $0.0001$, $0.001$, $0.01$, $0.1$, and $1$ were tested for each PyTorch ML task example. The value yielding the lowest $\Delta E_i^\mathrm{test,min}$ was employed in the performance evaluation (see Table S2 in the Supporting Information). For lMLP training the two most likely options according to the PyTorch ML task results were tested (see Table S4 in the Supporting Information).

\section{Results and Discussion}\label{sec:Results_and_Discussion}

\subsection{General Recommendations for CoRe Optimizer Hyperparameter Values}

A generally applicable set of CoRe optimizer hyperparameter values has been obtained from our benchmark on nine ML tasks including seven different models and six different data sets. The training processes span the entire range from learning on small mini-batches to full data set batch learning. Based on this benchmark we generally recommend the hyperparameter values $\beta_1^\mathrm{a}=0.7375$, $\beta_1^\mathrm{b}=0.8125$, $\beta_1^\mathrm{c}=250$, $\beta_2=0.99$, $\epsilon=10^{-8}$, $\eta_-=0.7375$, $\eta_+=1.2$, $s_\mathrm{min}=10^{-6}$, $s_\xi^0=10^{-3}$, $d_\chi=0.1$, and $t_\mathrm{hist}=250$. The number of frozen weights per group $n_\mathrm{frozen,\chi}$ can often be specified as a fraction of frozen weights per group $p_\mathrm{frozen,\chi}$. Well working values of $p_\mathrm{frozen,\chi}$ are typically in the interval between $0$ (without stability-plasticity balance) and about $10\%$. The maximal step size $s_\mathrm{max}$ is recommended to be $10^{-3}$ for mini-batch learning, $1$ for batch learning, and $10^{-2}$ for intermediate cases. $s_\mathrm{max}$ is the main hyperparameter like the learning rate $\gamma$ in many other optimizers.

\subsection{Optimizer Performance Evaluation for Diverse Machine Learning Tasks}

\begin{figure*}[tb!]
\centering
\includegraphics[width=\textwidth]{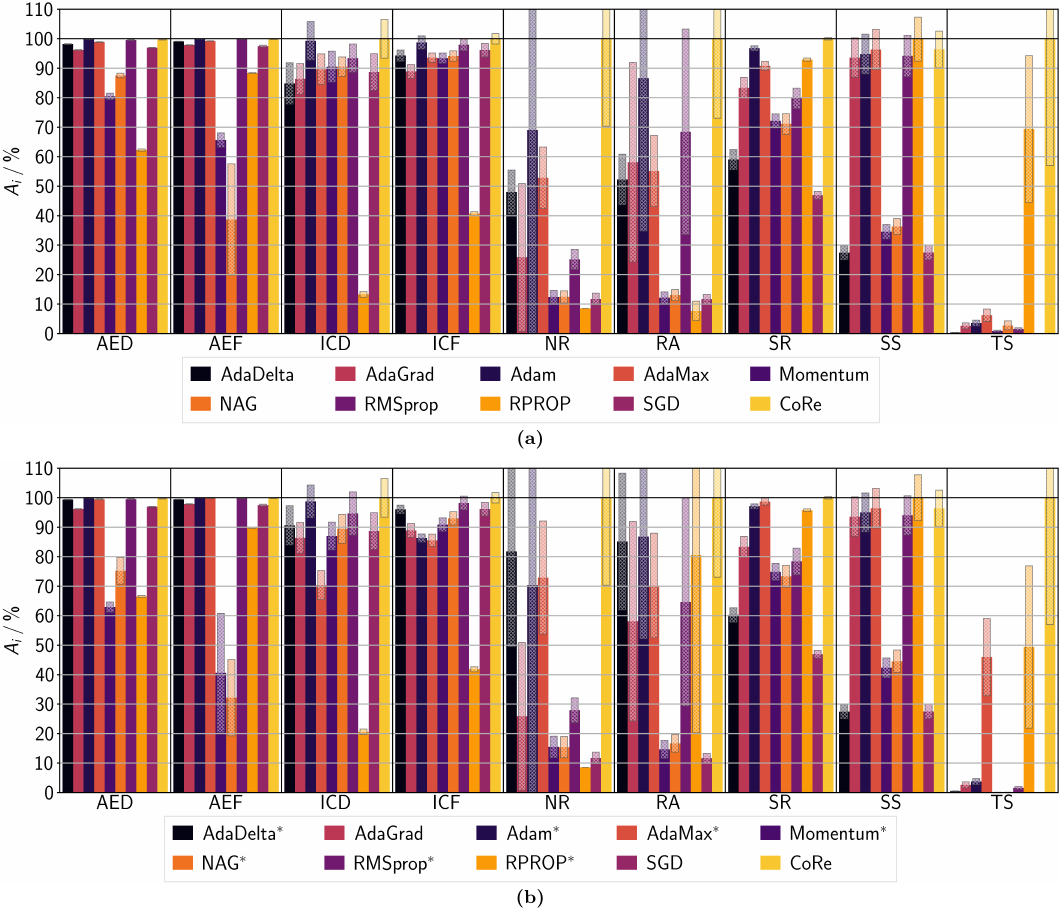}
\caption{Bar chart of the final accuracy scores $A_i$ (Equation (\ref{eq:A_i})) of various ML tasks $i$ trained by different optimizers. The uncertainty interval (Equation (\ref{eq:Delta_A_i})) is shown as cross-hatched bar around the upper edge of the bar which equals the mean of $A_i$ over 20 trainings (100 for NR and RA). $100\%$ corresponds to the highest obtained final accuracy of all optimizer specifications. The acronyms of the ML tasks are explained in Table \ref{tab:ML_tasks}. The optimizers are listed in the legends and are represented by different colors. The learning rate (maximal step size for RPROP, RPROP$^*$, and CoRe) was adjusted, while all other hyperparameters of the optimizers were set to \textbf{(a)} their general recommendation and \textbf{(b)} modified values. Exceptions are AdaGrad and SGD which do not include additional hyperparameters beyond the learning rate. The CoRe results are shown as reference in \textbf{(b)}.}\label{fig:final_accuracy}
\end{figure*}

To assess the performance of the CoRe optimizer in comparison to nine other optimizers with in total 16 different hyperparameter settings, relative accuracy scores for nine ML tasks were calculated for these optimizers (Figures \ref{fig:final_accuracy} (a) and (b)). For mini-batch learning on small batch sizes ($0.1\%$ for AED, AEF, ICD, and ICF) the popular Adam optimizer and our CoRe optimizer perform best, while especially RPROP yields poor accuracy because it cannot handle well stochastic gradient fluctuations. RPROP is intended for batch learning which becomes obvious by the high accuracy scores for SS and TS. For these ML tasks, RPROP and the CoRe optimizer achieve the highest accuracy scores. In the intermediate case, i.e., mini-batch learning with rather large batch sizes ($5\%$ for SR and $10\%$ for lMLP training (Figure \ref{fig:final_overall_accuracy_lMLP})), both Adam and RPROP perform well with Adam having a small advantage over RPROP. However, the CoRe optimizer outperforms both in this case.

\begin{table}[htb!]
\caption{Overview of the ML tasks and the respective data sets including their acronyms.}
\begin{center}
\begin{tabular}{lll}
\hline\vspace{-0.325cm}\\
 & ML task & Data set\vspace{0.075cm}\\
\hline\vspace{-0.325cm}\\
AED & auto-encoding & MNIST digits \cite{Deng2012} \\
AEF & auto-encoding & Fashion-MNIST \cite{Xiao2017} \\
ICD & image classification & MNIST digits \cite{Deng2012} \\
ICF & image classification & Fashion-MNIST \cite{Xiao2017} \\
NR  & naive reinforcement learning & cart-pole problem \cite{Barto1983} \\
RA  & reinforcement learning & cart-pole problem \cite{Barto1983} \\
    & (actor-critic) & \\
SR  & single image super-resolution & BSD300 \cite{Martin2001} \\
SS  & semi-supervised classification & Cora \cite{Sen2008} \\
TS  & time sequence prediction & sine waves \vspace{0.075cm}\\
\hline
\end{tabular}
\end{center}
\label{tab:ML_tasks}
\end{table}

\begin{figure}[htb!]
\centering
\includegraphics[width=\columnwidth]{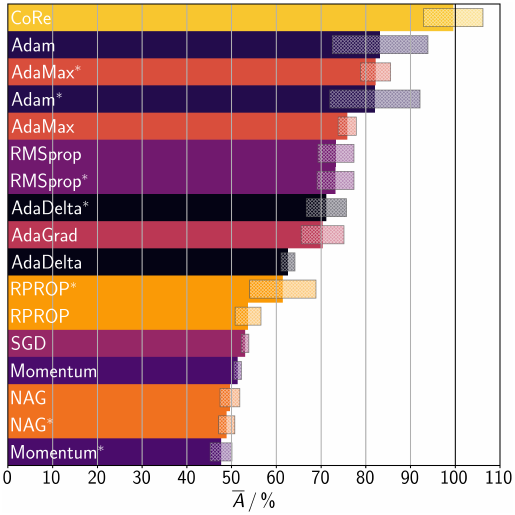}
\caption{Bar chart of the final accuracy score $\overline{A}$ (Equation (\ref{eq:A})) averaged over all ML tasks shown in Figures \ref{fig:final_accuracy} (a) and (b) for different optimizers. The uncertainty interval (Equation (\ref{eq:Delta_A})) is shown as cross-hatched bar around the right edge of the bar which equals the value of $\overline{A}$. A value of $100\%$ means that the optimizer achieves highest accuracy in every ML task. The bars are labeled and colored according to the respective optimizer.}\label{fig:final_overall_accuracy}
\end{figure}

Moreover, the learning speed and reliability of the CoRe optimizer in reinforcement learning (NR and RA) is also better than for the other optimizers (Figures \ref{fig:final_accuracy} (a) and (b)). RPROP is not able to learn the task in the maximal number of episodes for NR in any training. The CoRe optimizer's convergence speed of the mean test set errors for the other ML tasks is similar to Adam for mini-batch learning and similar to RPROP for batch learning (see Figures S1 to S7 and S9 in the Supporting Information).

In total, the CoRe optimizer achieves the highest final accuracy score in six tasks and lMLP training, Adam$^*$ in two tasks, and RPROP$^*$ in one task (Figures \ref{fig:final_accuracy} (a) and (b)). However, in the six cases where the CoRe optimizers performs best, the second best optimizer is always within the uncertainty interval of the CoRe optimizer's accuracy score. Still, there is no single optimizer which is always within the uncertainty interval. For example, Adam, RMSprop, RMSprop$^*$, and SGD are within the uncertainty interval for ML task ICF, only AdaMax$^*$ for SR, and AdaMax$^*$, RPROP, and RPROP$^*$ for TS, whereas the CoRe optimizer is always within the uncertainty interval of the best optimizer for the other three ML tasks. Hence, even if there is no clear dominance for individual ML tasks, the CoRe optimizer is among the best optimizers in all these ML tasks resulting in, on average, the best performance and the broadest applicability. Therefore, the CoRe optimizer is well-rounded and achieves the highest overall accuracy score (Figure \ref{fig:final_overall_accuracy}). The overall accuracy score of Adam is second highest, while those of AdaMax$^*$ and Adam$^*$ are almost equal to that of Adam. The uncertainty interval of the CoRe optimizer's overall accuracy score overlaps slightly with that of the Adam optimizer. We note that the uncertainty interval of the Adam optimizer's results is also the largest among all results.

In general, for the chosen set of ML tasks the optimizers which combine momentum and individually adapted learning rates (CoRe, Adam, and AdaMax) perform better than those which only apply individually adapted learning rates (RMSprop, AdaGrad, and AdaDelta) (Figure \ref{fig:final_overall_accuracy}). However, the differences among the CoRe optimizer, Adam, and AdaMax are larger than that of \text{RMSprop} and AdaMax. The final accuracy obtained by pure SGD is significantly worse than that of the aforementioned optimizers. However, for these nine ML tasks it is still slightly better than that of the optimizers which employ only momentum (Momentum and NAG). The overall accuracy of RPROP is in between those applying individually adapted learning rates and SGD for these ML tasks. However, this order is, of course, dependent on the fraction of mini-batch and batch learning ML tasks.

The best single model performances obtained by the CoRe optimizer are provided in Table S5 and Figures S10 (a) and (b) and S11 in the Supporting Information. For SS we can compare the final accuracy directly to the original work with $81.5\%$ correct test set classifications \cite{Kipf2017}. Due to training by the CoRe optimizer, the best graph convolutional network for SS achieves a test set classification accuracy of $84.2\%$.

\subsection{Performance Dependence on Hyperparameter Values}

The CoRe optimizer's hyperparameters were tuned on this set of ML tasks, while the general hyperparameter recommendations of PyTorch for the other optimizers were not based on this benchmark set. To provide a fair comparison, we also applied hyperparameter values for the other optimizers which were adjusted on this set of ML tasks. The adjusted hyperparameters of AdaDelta$^*$, AdaMax$^*$, Momentum$^*$, and RPROP$^*$ yielded an improvement of their overall accuracy scores (Figure \ref{fig:final_overall_accuracy}). However, the gain is not sufficient to reach the overall accuracy scores in the next better class of optimizers described in the last section. Therefore, the choice of the optimization algorithm is confirmed to be crucial for the final accuracy of the ML model. The highest overall accuracy scores of Adam, NAG, and RMSprop were obtained with their generally recommended hyperparameter values. The second best choices of the hyperparameters yielded very similar overall accuracy scores.

Another difference between the CoRe optimizer and the other optimizers was the application of a weight decay. However, Figures S13 and S14 in the Supporting Information show that the standard weight decay algorithm of Adam employed with four different hyperparameter values in general reduces the accuracy score for Adam. Only the weight decay algorithm of AdamW can lead to a small increase of the overall accuracy score. However, the gain is only a fraction of the overall accuracy score difference between the CoRe optimizer and the Adam optimizer. The weight decay of the CoRe optimizer only marginally affects the final accuracy on average (Figures S13 and S14 in the Supporting Information).

In the analysis of individual ML task performances, we note that RPROP and the CoRe optimizer show a slow convergence in the initial epochs of SS training (see Figure S6 in the Supporting Information). The reason is that large weight changes are required in the optimization and the initial step size $s_\xi^0$ is only set to $0.001$. Higher values of $s_\xi^0$ result in faster convergence to a similar final accuracy, with $s_\xi^0=0.1$ yielding a much faster convergence than obtained with Adam (see Figure S7 in the Supporting Information). However, this ML task is an extreme example with few weight updates to adjust $s_\xi^\tau$ in batch learning and the need of large weight changes. Still, as the final accuracy is the same and in most applications $s_\xi^\tau$ is fast adapted in a relatively small fraction of weight updates, the initialization of $s_\xi^0$ is in general noncritical.

Another edge case can be obtained for high maximal step size values $s_\mathrm{max}$ in the CoRe optimizer. While $s_\mathrm{max}=1$ yields a high final accuracy in TS training when early stopping is applied, the training can become unstable when continued (see Figure S8 in the Supporting Information). However, reducing $s_\mathrm{max}$ to $0.1$ already solves this issue (see Figure S9 in the Supporting Information).

\subsection{Optimizer Performance in Training Lifelong Machine Learning Potentials}

\begin{figure}[htb!]
\centering
\includegraphics[width=\columnwidth]{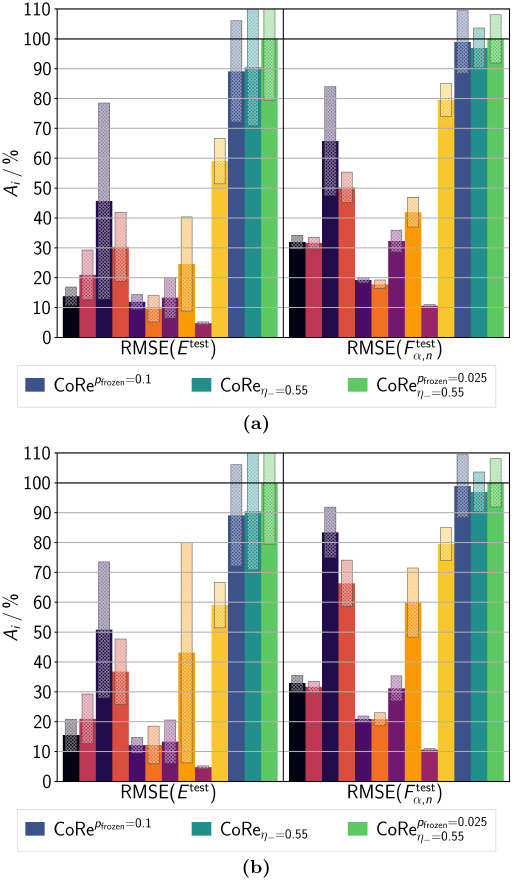}
\caption{Bar chart of the final accuracy scores $A_i$ (Equation (\ref{eq:A_i})) of energy and force prediction of lMLPs trained by different optimizers. The uncertainty interval (Equation (\ref{eq:Delta_A_i})) is shown as cross-hatched bar around the upper edge of the bar which equals the mean of $A_i$ over 20 trainings. $100\%$ corresponds to the highest obtained final accuracy of all optimizer specifications, i.e., the lowest RMSE in the prediction of energies or atomic force components. The colors of most optimizers are listed in the legends of Figures \ref{fig:final_accuracy} (a) and (b). The learning rate (maximal step size for RPROP and CoRe specifications) was adjusted, while all other hyperparameters of the optimizers were set to \textbf{(a)} their general recommendation and \textbf{(b)} modified values. Exceptions are AdaGrad and SGD which do not include additional hyperparameters beyond the learning rate. The CoRe results are shown as reference in \textbf{(b)}.}\label{fig:final_accuracy_lMLP}
\end{figure}

\begin{figure}[htb!]
\centering
\includegraphics[width=\columnwidth]{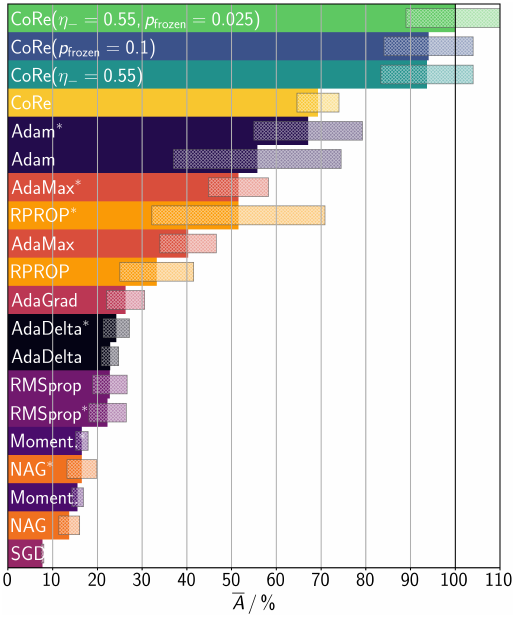}
\caption{Bar chart of the final accuracy score $\overline{A}$ (Equation (\ref{eq:A})) combining energy and force prediction of lMLPs for different optimizers. The uncertainty interval (Equation (\ref{eq:Delta_A})) is shown as cross-hatched bar around the right edge of the bar which equals the value of $\overline{A}$. A value of $100\%$ means that the optimizer achieves highest accuracy in energy and force prediction. The bars are labeled and colored according to the respective optimizer.}\label{fig:final_overall_accuracy_lMLP}
\end{figure}

In the training of lMLPs rather large fractions of training data ($10\%$) were employed in the loss function gradient calculation. In line with the results of the PyTorch ML task examples, this kind of training best suits the CoRe optimizer followed by Adam$^*$, Adam, Adamax$^*$, and RPROP$^*$ (Figures \ref{fig:final_accuracy_lMLP} (a) and (b) and \ref{fig:final_overall_accuracy_lMLP}). Moreover, the general trend is confirmed that adaptive and momentum based optimizers perform best, while only adaptive optimizers still yield better results than only momentum based optimizers. In contrast to the PyTorch ML task examples, where the stability-plasticity balance of the CoRe optimizer with $p_\mathrm{frozen}$ around $0.025$ can only marginally improve the accuracy scores for AED, AEF, and SR and worsens the final accuracy for ICD and ICF (see Figure S12 in the Supporting Information), the lMLP training largely benefits from the stability-plasticity balance with $p_\mathrm{frozen}=0.1$. We note that tuning $p_\mathrm{frozen}$, in addition to the maximal step size, extends the hyperparameter optimization capability for $\mathrm{CoRe}^{p_\mathrm{frozen}=0.1}$ compared to CoRe and all other optimizers, for which only the maximal step size/learning rate was adjusted while all other hyperparameters were taken from the PyTorch ML task example results. This higher degree of freedom can also contribute to the optimization performance. However, a significant performance improvement was not obtained for any other hyperparameter tuning with the exception of tuning $\eta_-$.

\begin{figure}[htb!]
\centering
\includegraphics[width=\columnwidth]{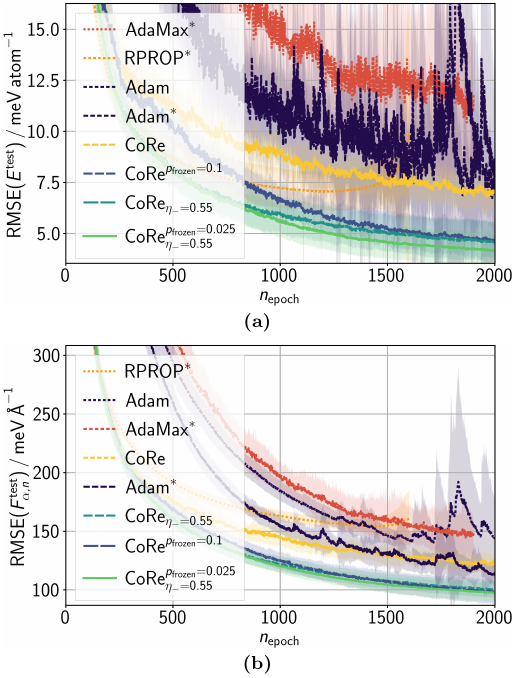}
\caption{Test set RMSEs of \textbf{(a)} energy $E^\mathrm{test}$ and \textbf{(b)} atomic force components $F_{\alpha,n}^\mathrm{test}$ as a function of the training epoch $n_\mathrm{epoch}$ for the lMLP compared to the DFT reference. The results are shown for the eight optimizers yielding highest final accuracy. The less often a line is broken, the lower is the final error. Uncertainty intervals are shown in pale color of the respective line.}\label{fig:convergence_lMLP_E_F}
\end{figure}

Moreover, the stability-plasticity balance smoothens the training convergence as shown in the test set root mean square errors (RMSEs) of energies and atomic force components as a function of the training epochs (Figures \ref{fig:convergence_lMLP_E_F} (a) and (b)). The CoRe optimizer yields smoother convergence than Adam, which is beneficial, for example, in lifelong machine learning where the lMLP needs to be ready for application in every training stage. The accuracy scores in Figures \ref{fig:final_accuracy_lMLP} (a) and (b) and \ref{fig:final_overall_accuracy_lMLP} take into account the convergence smoothness (see Section \ref{sec:Computational_Details}) in contrast to the accuracy scores in Figures S15 and S16 in the Supporting Information which are only based on the individual lMLP early stopping results. The latter is beneficial for the Adam results but still the CoRe optimizer with stability-plasticity balance outperforms Adam. The convergence speed is also higher for the CoRe optimizer than for Adam. This observation is in line with the convergence of the SR ML task (see Figure S5 in the Supporting Information) which also represents a training case between mini-batch and batch learning. To demonstrate the benefit of more stabilized learning in lMLP training, we additionally decreased the $\eta_-$ value which smoothens and improves the training process similarly. Both, a large $p_\mathrm{frozen}$ and a small $\eta_-$, lead also to a better interplay with the lifelong adaptive data selection. However, this interplay is only a minor factor of the large accuracy score improvement since the improvement is similar in training with random data selection (see Figure S17 in the Supporting Information). Lifelong adaptive data selection increases the final accuracy in general. In conclusion, a very smooth convergence is desired in lMLP training making a smaller $\eta_-$ value beneficial. However, the final accuracy and convergence speed and smoothness are already higher than those of other state-of-the-art optimizers when the generally recommended hyperparameter values with a stability-plasticity balance enabled by $p_\mathrm{frozen}$ are applied.

In comparison to our previous work, where the best 10 of 20 lMLPs yielded $\mathrm{RMSE}(E^\mathrm{test})$ and $\mathrm{RMSE}(F_{\alpha,n}^\mathrm{test})$ to be $(4.5\pm0.6)\,\mathrm{meV\,atom}^{-1}$ and $(116\pm4)\,\mathrm{meV}\,\text{\AA}^{-1}$ after 2000 training epochs with the CoRe optimizer, the generally recommended hyperparameters of this work in combination with $p_\mathrm{frozen}=0.1$ ($\mathrm{CoRe}^{p_\mathrm{frozen}=0.1}$) improved the accuracy to $(4.1\pm0.7)\,\mathrm{meV\,atom}^{-1}$ and $(90\pm5)\,\mathrm{meV}\,\text{\AA}^{-1}$. With an adjusted $\eta_-$ value for even smoother training ($\mathrm{CoRe}_{\eta_-=0.55}^{p_\mathrm{frozen}=0.025}$) the respective test set RMSE values decreased to only $(3.4\pm0.4)\,\mathrm{meV\,atom}^{-1}$ and $(92\pm4)\,\mathrm{meV}\,\text{\AA}^{-1}$.

Finally, the comparison of computation time for training with Adam and the CoRe optimizer shows that not only the final accuracy but also the accuracy-cost ratio of the CoRe optimizer is better than that of Adam. For comparison of multiple trainings with the lMLP software, the time fraction of model fitting in the entire training process (including initialization, descriptor calculation, model fitting (about $87\%$), final prediction, and finalization) is calculated to reduce the influence of different computers and computation loads. The resulting speed is the same within the uncertainty interval for Adam and the CoRe optimizer. The additional operations in the CoRe optimizer algorithm cause only little increase of computational cost which is not significant in comparison to the cost for evaluating the loss function gradient. For the presented lMLP example, an optimizer step requires less than $0.2\%$ of the time needed for a loss function gradient calculation. Since the CoRe optimizer requires only the loss function gradient as input like Adam and the other optimizers, the computation time per training epoch is similar for all optimizers.

\section{Conclusion}\label{sec:Conclusion}

The CoRe optimizer combines Adam-like and RPROP-like weight-specific learning rate adaption. Moreover, in the CoRe optimizer step-dependent decay rates are employed in the calculation of Adam-like gradient moving averages, which are the basis of the RPROP-like step size updates. Its weight decay depends on the absolute weight update and an optional stability-plasticity balance based on a weight importance score can be applied. In this way, the CoRe optimizer combines the high performance of the Adam optimizer in small mini-batch learning and that of RPROP in full data set batch learning, while it is superior to both in intermediate cases. With the general hyperparameter recommendation obtained in this work based on diverse ML tasks, the CoRe optimizer is a well-rounded all-in-one solution with broad applicability and high convergence speed and final accuracy on-par and beyond state-of-the-art first-order gradient-based optimizers.

The performance evaluation has further confirmed a general advantage for optimizers which combine momentum and individually adapted learning rates in terms of convergence speed and final accuracy compared to optimizers which are only adaptive or momentum based or none of these. Moreover, adaptive and/or momentum based methods need only marginally more computation time than simple SGD which is negligible compared to the time required for loss function gradient calculation.

Besides the general CoRe optimizer hyperparameter recommendation, only the maximal step size $s_\mathrm{max}$ needs to be set depending on the fluctuations in the gradient calculation which can be estimated easily based on the application of mini-batch ($0.001$) or batch learning ($1$) or intermediate cases ($0.01$). Additionally, the stability-plasticity balance can be enabled by the hyperparameter $p_\mathrm{frozen}$. It can achieve smoother training convergence to even higher final accuracy yielding a large improvement in the example of lMLP training. We note that hyperparameter fine-tuning for individual ML tasks can, of course, improve the performance to some degree for all optimizers but comes with the drawback of being very time consuming. 

\section*{Code Availability}

The CoRe optimizer software is available on GitHub (\url{https://github.com/ReiherGroup/CoRe_optimizer}) and PyPI (\url{https://pypi.org/project/core-optimizer}‌).‌

\hfill\newpage

\section*{Acknowledgement}

This work was supported by an ETH Zurich Postdoctoral Fellowship.

\section*{Supporting Information}

Optimizer hyperparameters including adjusted learning rates or maximal step sizes; optimizer performance comparison for ML tasks AED, AEF, ICD, ICF, SR, SS, and TS; performance of best single models trained by the CoRe optimizer; final accuracy for ML tasks AED, AEF, ICD, ICF, and SR trained by the Core optimizer with stability-plasticity balance; final accuracy obtained by the Adam optimizer with weight decay; optimizer performance comparison for lMLPs applying early stopping; final accuracy for lMLPs trained by the CoRe optimizer applying random data selection or lifelong adaptive data selection (PDF file available free of charge at DOI 10.1088/2632-2153/ad1f76).

%

\end{document}